# Incorporating Image Gradients as Secondary Input Associated with Input Image to Improve the Performance of the CNN Model

Vijay Pandey[1, *], Shashi Bhushan Jha[2]

[1]Department of Computer Science Engineering, IIT Kharagpur, India

[2]Department of Electrical Engineering and Computer Science, Embry-Riddle Aeronautical University, Daytona Beach, FL 32114, USA

E-mail ID: vijayiitkgp13@gmail.com, jhas1@my.erau.edu

*Corresponding author (email: vijayiitkgp13@gmail.com)

## ABSTRACT

CNN is very popular neural network architecture in modern days. It is primarily most used tool for vision related task to extract the important features from the given image. Moreover, CNN works as a filter to extract the important features using convolutional operation in distinct layers. In existing CNN architectures, to train the network on given input, only single form of given input is fed to the network. In this paper, new architecture has been proposed where given input is passed in more than one form to the network simultaneously by sharing the layers with both forms of input. We incorporate image gradient as second form of the input associated with the original input image and allowing both inputs to flow in the network using same number of parameters to improve the performance of the model for better generalization. The results of the proposed CNN architecture, applying on diverse set of datasets such as MNIST, CIFAR10 and CIFAR100 show superior result compared to the benchmark CNN architecture considering inputs in single form.

## 1. INTRODUCTION

In the past, various research works have been carried out in neural network field. In 1989, CNN (convolutional neural network) acquired the attention that paved the new way for research in the field of computer vision. In the beginning, LexNet, AlexNet were developed where concept of CNN was applied. Then in the following years, innovation occurred, however, most of the researchers stressed on targets to make network deeper and deeper. After few more years, when deeper networks attained popularity with enhanced computational power and increased data size then some problems were noticed. These problems are vanishing gradient and exploding gradient problem on deeper networks. Therefore, residual neural networks were developed to tackle such problem. In the current time, to reduce the weight size as well as the computational complexity, the inception model has been proposed for the channel-wise training. Later, several regularization techniques have been evolved that includes but not limited to, dropout, Gaussian regularization, adding penalty term in loss function, drop connect, batch normalization. Various new tricks like, dropout, connection dropout (Srivastava, et. al., 2014), block dropout, techniques have been used for the better generalisation of the data. In fact, batch normalization (Ioffe & Szegedy, 2015; Santurkar et al., 2018) is used for speed up training in gradient descent-based optimization setup. Numerous researches have been carried out for activation function, loss function, and batch normalization. Data augmentation is one of the areas where researchers focused on to improve the generalization power of the image. To enhance the model efficiency, an image augmentation technique exists to add image by scaling, rotating, and shearing. In fact, to improve the performance, this method facilitates to train the model efficiently considering the same training data. Although deep CNNs were widely recognized by the researchers in the past for processing image, text, speech and video. CNNs have attracted much attention in experimental evaluation. The existing researches focus on the architecture and on the ways to improve the training. Most of them does not focus on the form of input. They assume that, input is in pixel value form. It may be in 1D vector as in FCNN input, or may be in 2D, or 3D (including channel information), as in CNN. The current literature stressed that there are lot of scope to enhance existing optimization techniques to learn deep CNN architectures (He et. al., 2016; Huang et. al., 2016; Szegedy et. al., 2017). Particularly, the fundamental theory of CNNs still needs improvement. Although, the existing CNN model solves the problem effectively, there is a lack of understanding in how the model works fundamentally. Therefore, the fundamental principles of CNNs requires more investigation. In the meantime, to enrich the design of CNN requires much attention to explore in leveraging natural visual perception mechanism. This research can be useful for a depth understanding of CNNs and it assists for further studies and application advancements in the area of CNNs.

In this research, incorporating image gradients as input is to improve the accuracy and behind the improvement of accuracy, it has some theoretical significance. CNN works on the principal of feature extraction where it extracts the important features of the image that assists to generalize the classification task over the unseen images. This technique has some limitations that it is dependent on the inputs. From given images, it extracts the features and based on those features it helps in classifying the images. Here if we give the input in the different form along with the original input then it facilitates in extracting the features from two different images. Moreover, weights will be properly trained because it has now to take charge of two different forms of the same input hence better generalization. For transformation, this study focuses on the image transformation instead of depending on weight that saves the number of parameters and layers. Given an extra input has the significant impact on the model training, it facilitates to see the input in different aspects at the same time and the model learns from both the aspects. In this paper, it is noticed that if the new added image aspect is not good then it decreases the model performance significantly. In addition, if the added aspects are good enough then it improves the model performance considerably. Here on the concept of image augmentation, we are adding the different form of image, which is image gradient. In fact, both forms of input are being trained simultaneously rather than separately as in image augmentation technique.

## 2. RELATED WORK

In the past few decades many variants of CNN architectures has been proposed by several researchers. However, the fundamental mechanisms of CNNs are somewhat similar. In particular, CNN established as a deep learning architecture motivated by visual perception mechanism of the living beings. (Hubel & Wiesel, 1968) studied the receptive fields and functional architecture of animal visual cortex and further discovered that cells are accountable to detect light. Motivated by this study, (Fukushima, Miyake, & Ito, 1983) developed a Necognitron to recognize a visual pattern with the help of neural network model, the predecessor of CNN. Further, (Yann LeCun et al., 1990) presented a handwritten digit recognition problem using back-propagation network that established a modern model of CNN, and this research improved by (Y LeCun et al., 1998). In this study, authors considered multi-layer neural networks, namely, LeNet-5, to train the model using back propagation method with the aim to classify the handwritten digits. In another research, (Zhang, Itoh, Tanida, & Ichioka, 1990) proposed a shift-invariant interconnection with neural network to determine characters from the given image. During that period, it was difficult to tackle complex problems due to dearth of huge training data and high-performance computing system. However, several algorithms have been proposed after 2006 to tackle such difficulties in deep training deep CNNs (Niu & Suen, 2012; Russakovsky et al., 2015; Szegedy et al., 2015; Zeiler et al., 2014).

Most substantial work, a classic CNN architecture was developed by (Russakovsky et al., 2015) and it illustrated substantial improvements on previously published methods with respect to image classification task. In this study, the architecture of the model such as AlexNet (Russakovsky et al., 2015) is like LeNet-5, however, considering a deeper structure. Afterwards, several researches were proposed for improving the performances of the AlexNet model. In which, most notably, VGGNet (Simonyan & Zisserman, 2014), ZFNet (Zeiler et al., 2014), ResNet (He et al., 2016), and GoogleNet (Szegedy et al., 2015) are the popular four typical works. During this period, the main aim was to study the networks deeper such as ResNet that received the champion of ILSVRC in 2015 and in fact, ResNet is 20 times, and 8 times more depth than AlexNet, and VGGNeT, respectively. By growing more depth, the network can estimate the target function accurately considering enhanced nonlinearity and receive improved feature representation. Nonetheless, there is a drawback to increase the complexity of the model, it makes the network hard for optimization and leads to overfitting. However, in the past decade, many researches have dealt with distinct aspects of CNN involving convolutional and pooling layer, loss function, activation fiction, optimization and regularization.

With respect to network architecture, *Tiled convolution, Dilated convolution, Transposed convolution, Network in network, residual network, Inception module* are the widely recognized network. Similarly, with respect to regularization of the model, *Lp-norm* regularization*, DropConnect*, and *Dropout* are broadly known regularization technique. For the better computational efficiency of CNNs, *FFT (Fast Fourier Transform), structured transform, Weight Compression, Low precision, Sparse convolution* are extensively used.

## 3. ARCHITECTURE

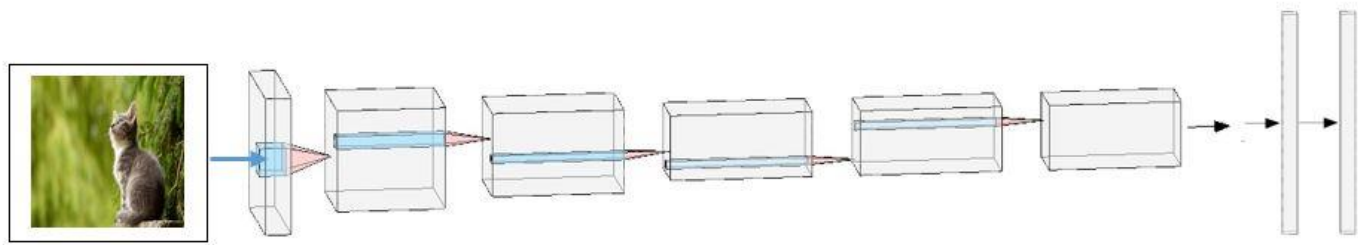

Fig.1. Traditional CNN architecture

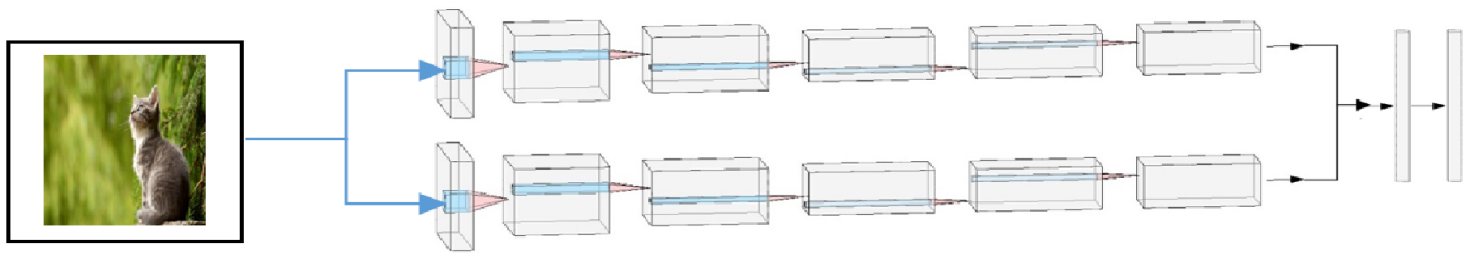

Fig. 2. Two paths in CNN architecture, one path is the copy of another. In the first path, original image in pixel form is passed and in the second path, image gradient input form is passed. Layers are shared between two branches. Before fully connected layer output of the both paths are added and then passed to further layers.

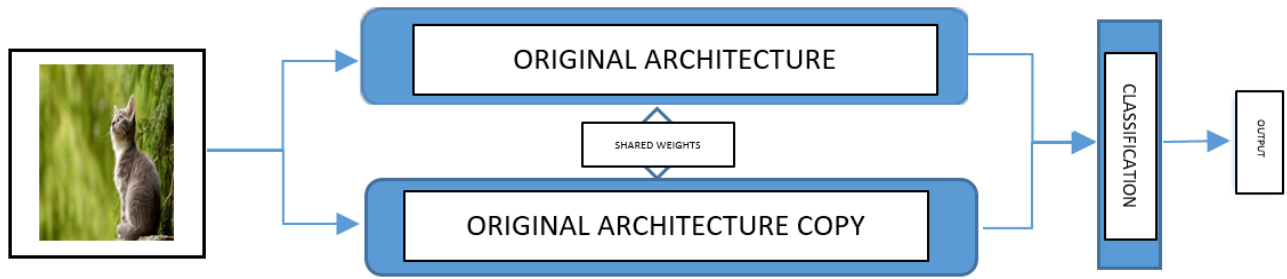

Fig. 3. Illustration of generalized architecture of Fig. 2.

In this section, architecture of CNN is explained. Initially, the traditional CNN architecture is presented that can be seen in Fig. 1. Furthermore, the proposed architecture of this research is illustrated in Fig. 2. In this architecture, first horizontal and vertical gradients, $dx$ and $dy$, respectively of each pixel position in image is calculated. Afterwards, one copy of the original image is created and for every pixel position, the value of the pixel with $f(dx, dy)$ is replaced. The new obtained image along with the original image is passed to the network in parallel. All feature extraction convolutional layers before dense layers are shared for both inputs. Adding extra relevant input facilitates in learning the weights in better way without adding extra parameters. Two paths in CNN architecture, one path is the copy of another. Before classification layer output of the last convolutional block of both branches are added together and are passed to the fully connected layer. In last, the illustration of generalized architecture of Fig. 2. is shown in Fig. 3.

## 4. WORKING PHILOSPHY OF PROPOSED ARCHITECTURE

This section focuses on working philosophy of proposed architecture. For instance, if a task is provided to a human in different-different forms then concept can be learnt in a better way. It learns what the meaningful pattern is provided in various forms, which leads these forms to accomplish the same target. In addition, a human brain system does not learn pattern efficiently from a single source, but they learn more robust pattern using different source of information. Thus, providing new input form consisting image gradient along with pixel form of image gives the same effect as in the above example. Moreover, a source of extra diverse information helps in generalizing and finding the insight in the image better than learning form a single source of image information. With respect to rich representation concept, it represents the concept to be learned in proper manner. If rich representation is fed into the network, then it learns better. Rest how network is extracting the feature depends on hyper-parameters and network architecture. In further subsections, discussion will be on various aspects of the proposed architecture.

### 4.1. Providing image gradient as parallel input to pixel value input

Idea of providing extra input came from available literatures. (Simard et al., 2003) suggests that elastic distortions in data set is helpful for training. Corrupting the training examples with noise from known distributions within the exponential family gives robust training for the network (Matane et al., 2013), this study mentioned that if you add some noise in the data from outside then it becomes very well trained on the given data and weight are updated well for this process. (Bishop, 1995; Jiang et al., 2009; Matsuoka, 1992; Murray & Edwards, 1994; Wang & Principe, 1999) all the mentioned literatures suggest, if some substantial feature engineering is considered on given inputs such as adding noise, corrupting the features, data distortions, and feature deletion then it assists network to learn more robust and provides better generalization on unseen data.

By adding extra input form as image gradient, the overall goal of task to classify image remains same. Image gradients input acts as regularizer by providing supervised pre-training. Supervised pre-training because prior knowledge that image gradient is one of the key feature of image is passed to network. Due to the same, it works as a good initialization method by making the use of rich representation of image in form of image gradients (Murray & Edwards, 1994; Reed et al., 1995).

**4.2. Weight sharing**

As convolutional network works on the philosophy of weight sharing. Therefore, weight sharing facilitates in knowing the common pattern across the given input space where the weight is applied. Sharing weight means same set of parameters are applied to each sequential branch and each branch regularizes each other. In convolutional neural network, suppose there is an input 2D image $I$, where we are applying a convolution operation with a filter $F$ of kernel size 3*3 in layer $L$, so the total number of weights are nine for that layer (excluding bias and assuming only a single filter is used). Therefore, this weight is applied over entire image. On each patch of 3*3 with stride 1, this weight is applied. It will try to learn the common pattern across that patches. Thus, if these patches are as much diverse in representing the same concept then these weights will be trained much well, because they will try to learn features that are more robust without being overfit in case of less diverse input space. It will work as a regularizer, which will help in leaning the weights properly (Han et al., 2016; Huang et al., 2013).

**4.3. Multi input forms single output (multitasking)**

Multi-tasking works on the concept of the weight sharing, where there is one input matrix and multiple tasks to predict. Information needed to predict those tasks is represented by that single input matrix. For instance, in face detection task, detecting face, eye, ear, and nose each can be considered as individual task. Information to predict these tasks to detect face, eye, ear, and nose are represented using the single face image matrix. Thus, weights are trained in more robust way to focus on every task at the same time rather than focusing on only a single task at given time. This process of multitasking makes network learn well and generalize well by learning multiple tasks at the same time (Thung & Wee, 2018; Y. Zhang & Yang, 2017; Y. Zhang & Yeung, 2014). In the proposed architecture, the same concept as of multiple tasking is applied with reversing the direction of flow. In traditional multiple tasking, there is single input matrix and multiple task, however, in our approach, there are multiple input matrix and single task. Consequently, weight sharing is to capture multiple representations of information to produce single task. Hence, it will give the same effect as multitasking for better learning of the network.

**4.4. Feeding diverse data for classification**

In proposed architecture, feature maps of last convolutional block of both paths are added together before passing it to the classification layer. This facilitates to train the classifier by consisting enrich and diverse representation of the image. Therefore, classification layer weights learn better. Moreover, it can be concatenated as well but concatenation of layers will increase the parameter size, which is not intended here. In addition, to train the classification layer there can be three different potential ways. (1) Adding the two feature maps, which will work as relevant noise to each other. (2) Training the weights of the classification layer in shared manner as happening in convolutional network. In this case, output of both paths will be added before softmax layer.

Adding diverse input representation assists in better generalization as mentioned in various previous researches. As in the HOG based face detection (Dalal & Triggs, n.d.), histogram of face image gradients is used to classify the face in a given image. Therefore, in proposed architecture, image gradient feature map along with image feature map is passing to the classifier that makes pattern more robust for better classification result.

**4.5. Data representation, rich diverse prior**

Training of the model depends on the quality of the data, how much the data is enriched. If the data will be enriched, then it can represent the knowledge in a better way. It should include the information that can be encoded by the neural network layers for better prediction. Each layer in the neural networks tries to learn the filters to extract the important features that assists in classification task of the model. In this study, the data is represented in the diverse form of single input. The gradient input has been considered in this research, can be seen as a feature map that has been passed through two filters: first, to calculate the gradient, and second, to compute the sum of that gradient for pixel's *x* and *y* position. This technique decreases the number of layers that means it reduces the complexity of the network and number of parameters as well. Moreover, the key advantage of proposed approach is to represent the data in form of different features. Therefore, if the proposed technique passes the extra relevant features to the network then it encodes the concept information substantially and facilitates in learning the model.

### 4.6. Providing regularisation effect

In the paper (J. Zhang et al., 2020), it is mentioned, "Regularization influences the encoding of relative features with prediction task in neural network". In proposed architecture, weight sharing, adding extra relevant image feature (noise) with input and applying multitasking concept provides the regularization effect in network training.

### 4.7. Reducing overfitting by providing relative information

One of the key cause for overfitting is to introduce irrelative information with the target (J. Zhang et al., 2020). If we provide irrelative information then it causes the overfitting because the network will be trained based on irrelevant information, which might be missing in the test data. On the other hand, if we offer the relative information with the target, then the generalization can be achieved in a better way. Thus, this research provides image gradient as another input to the network for more relevant information that facilitates the model in reducing the overfitting.

### 4.8. Similarity with natural visual perception mechanism

This section demonstrates the similarity of this study with natural visual perception mechanism. Human visual perception mechanism receives input and then start processing it using various mechanisms. Modern convolutional neural networks (CNNs) work in similar fashion, where it takes image and process it using convolutional layer, activation layer, pooling layer and then classification layer. In this paper, two forms of same input are considered at the same time. It is, as if one person is seeing one scene in two different form at the same time. One is in the pixel value form and another is the summation of the horizontal and vertical direction gradients of given image. As in human visual perception mechanism, if multiple forms of same input are received to represent the same target then resources are not increased rather available resources are trained more robustly. The same principle has been considered for this study; thus, no extra parameters are increased. In fact, the proposed approach is to update the weights in proper way with available resources. However, what sort of extra inputs are received by eyes that also matters. In this paper, it has been discovered that the different input forms have different impact on performance.

## 5. EXPERIMENTS

This section illustrates the experimental results of the proposed architecture. To perform the experiments, three most popular datasets in the field of computer vision are considered, namely, MNIST, CIFAR10 and CIFAR100. First horizontal and vertical gradient *dx* and *dy* respectively of each pixel position of image are calculated. For each dataset, two experiments are carried out. In first experiment, one copy of the original image is created and for every pixel position, the value of the pixel is replaced with *(dy + dx)*. In addition, the new obtained image along with original image are provided to the network in parallel. In second, only original image is fed to the network.

The main aim of all the experiments is to show that adding image gradient as an extra input source improves the performance of the model significantly. Particularly, this research mainly focuses on the fact that in the given architecture if image gradient is added then it increases the generalization ability of the original network without increasing the parameter.

The network architecture detail of second type of experiment in which only original image is provided as input to the network is shown in Table 1.

| Neural Networks For Experiment ||||
|---|---|---|---|
| Layer | MNIST | CIFAR10 | CIFAR100 |
| conv2d_1 | [3*3, 4], Padding 1 | [3*3, 16], Padding 1 ||
| activation_1 | RELU |||
| max_pooling2d_1 | 2*2 MAX, Stride 2 |||
| dropout_1 | 0.2 |||
| batch_normalization_1 | Normalize activations of previous layer at each batch |||
| conv2d_2 | NA | [3*3, 32], Padding 1 ||
| activation_2 | NA | RELU ||
| dropout_2 | NA | 0.2 ||
| batch_normalization_2 | NA | Normalize activations of previous layer at each batch ||
| conv2d_3 | NA | [3*3, 64], Padding 1 ||
| activation_3 | NA | RELU ||
| dropout_3 | NA | 0.2 ||
| batch_normalization_3 | NA | Normalize activations of previous layer at each batch ||
| flatten_1 | convet matrix into vector |||
| dense_1 | 64 | 128 ||
| dense_2 | 10 | 128 ||
| dense_3 | NA | 10 | 100 |
| softmax | OUTPUT CLASS PROBABILITIES |||

Table 1: Architecture detail of experiment where only original image is passed as input to the network.

It can be noticed from the given architecture in Table 1 that for MNIST data set, one convolution block, and for rest two datasets, three convolution blocks have been used. BatchNoramlization layer and dropout have been used to speed up the training and provide regularizing effect, respectively.

In Table 2, the architecture detail of first type of experiment is given where original image along with its gradient image are provided as parallel input. In fact, in Table 2, architecture is almost same as the architecture in Table 1 with few changes such as addition of *lambda* layer for calculation of image gradient, passing two diverse form of inputs in parallel having shared layers. Introducing *Add* layer that combines two parallel output vectors and passes combined input to dense layers. In further subsections, experiment on each above-mentioned dataset are conducted with their analysis.

| Neural Network For Experiment | | | | |
|---|---|---|---|---|
| Layer | | MNIST | CIFAR10 | CIFAR100 |
| Lambda (calculate image gradient) | Input Matrix | calculate image gradient and pass to subsequent layers parallely with original input | | |
| conv2d_1 | | [3*3, 4], Padding 1 | [3*3, 16], Padding 1 | |
| activation_1 | | RELU | | |
| max_pooling2d_1 | | 2*2 MAX, Stride 2 | | |
| dropout_1 | | 0.2 | | |
| batch_normalization_1 | | Normalize activations of previous layer at each batch | | |
| conv2d_2 | | NA | [3*3, 32], Padding 1 | |
| activation_2 | | NA | RELU | |
| dropout_2 | | NA | 0.2 | |
| batch_normalization_2 | | NA | Normalize activations of previous layer at each batch | |
| conv2d_3 | | NA | [3*3, 64], Padding 1 | |
| activation_3 | | NA | RELU | |
| dropout_3 | | NA | 0.2 | |
| batch_normalization_3 | | NA | Normalize activations of previous layer at each batch | |
| flatten_1 | | convert matrix into vector | | |
| Image Gradient Output | Image output | Two distinct feauture maps from two parallel distinct inputs | | |
| ADD | | Adding Both Feature Maps | | |
| dense_1 | | 64 | 128 | |
| dense_2 | | 10 | 128 | |
| dense_3 | | NA | 10 | 100 |
| softmax | | OUTPUT CLASS PROBABILITIES | | |

Table 2: Network architecture for experiment where image gradient (calculated using *lambda* layer) and original image are passed in parallel. All layers after *lambda* layer and before *ADD* layer are shared. After *ADD* layer both parallel branch outputs are added and are passed as single input to further layers.

### 5.1. MNIST

This is a dataset of 60,000, 28x28-grayscale images of the 10 digits along with a test set of 10,000 images. With this dataset, the experiments on one of the benchmark architectures and the proposed architecture are executed; the proposed architecture shows superior performances on both the train and test sets.

For test set performance curve is depicted in Fig. 4. In Fig. 4, one can clearly notice that the proposed network architecture performs significantly better in comparison to the benchmark architecture throughout all the epochs. Similarly, the performances of train set can also be observed in Fig. 5.

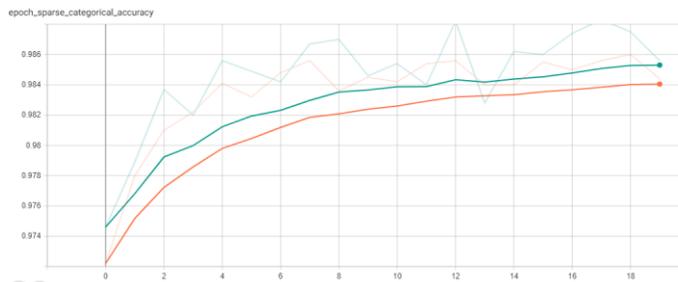 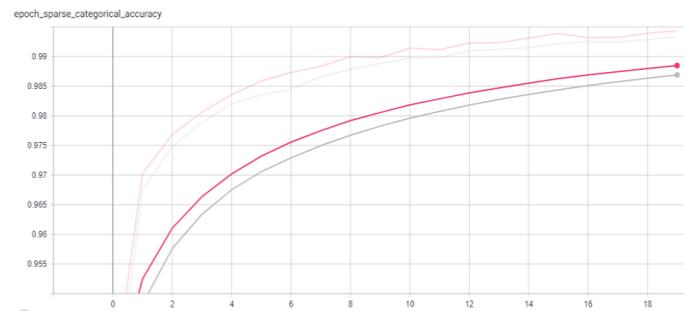

Fig. 4. Test accuracy curve, orange curve for benchmark architecture and green curve for proposed architecture. Epoch for x- axis, accuracy for y-axis.

Fig. 5. Train accuracy curve, grey curve for benchmark and pink curve for proposed one. Epoch for x-axis, accuracy for y-axis.

### 5.2. CIFAR10

CIFAR10 is a dataset of 50,000, 32x32 for colour-training images and 10,000 for test images, labelled over 10 categories. Considering this dataset, the experiments on benchmark architecture of CNN and the proposed architecture are carried out, the results determined by the proposed architecture illustrates better performance on both the train and test set. In fact, the test set performance curve is depicted in Fig. 6. In Fig. 6, one can clearly observe that the proposed network architecture performs superior in comparison to benchmark architecture throughout all the epochs. Similarly, the performances of train set are presented in Fig. 7.

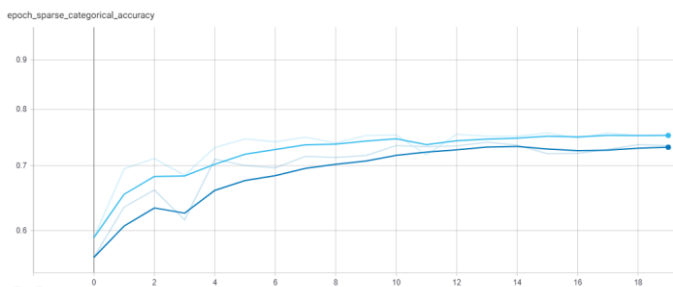 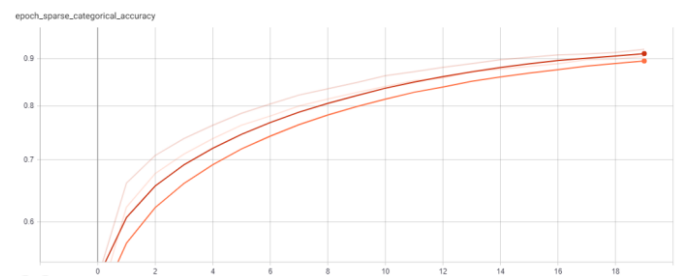

Fig. 6. Test accuracy curve, blue curve for benchmark architecture and light blue curve for proposed architecture. Epoch for x-axis, accuracy for y-axis.

Fig. 7. Train accuracy curve, orange curve for benchmark and red curve for proposed one. Epoch for x-axis, accuracy for y-axis.

### 5.3. CIFR100

This is a dataset of 50,000, 32x32 colour-training images and 10,000 test images, labelled over 100 fine-grained classes that are grouped into 20 coarse-grained classes.

In this section, the experiments are performed similar as shown in section 5.2 with different datasets. The experiments are executed on the benchmark architecture of CNN and the proposed architecture of this research and the results of both the architectures are determined. In fact, the proposed architecture shows better performance on both train and test set compared with the benchmark architecture. The test set performance curve is depicted in Fig. 8. In

Fig. 8, one can note that results obtained by the proposed network architecture are superior than the solutions determined by benchmark architecture throughout all the epochs. Moreover, the performance of train set is shown in Fig. 9.

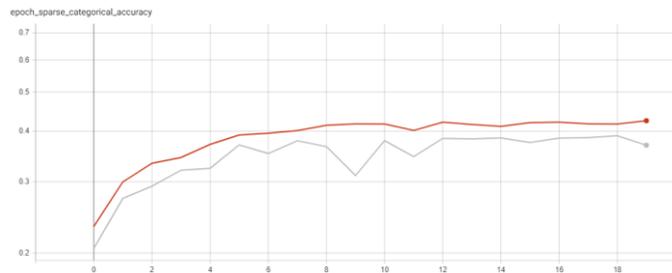

Fig. 8. Test accuracy curve, grey curve for benchmark architecture and red curve for proposed architecture. Epoch for x-axis, accuracy for y-axis.

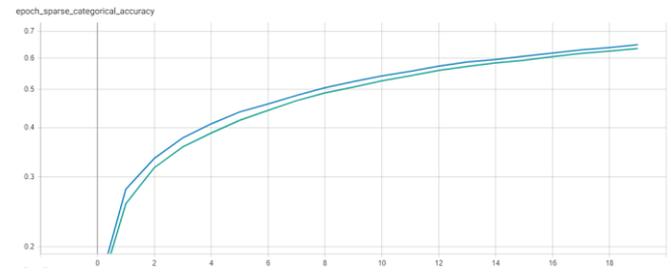

Fig. 9. Train accuracy curve, green curve for benchmark and blue curve for proposed one. Epoch for x-axis, accuracy for y-axis.

## 6. DISCUSSION

From the above results, it can be concluded that the proposed architecture evidently shows significant improvement over the benchmark architecture. In addition, it provides the regularizing effect in the network training. Another advantage of the network is, without adding extra layers or additional parameters model attains better generalization. Further researches can be carried out considering the proposed architecture by adding more diverse inputs like gradient of image. Moreover, one can add texture of image, sober edge of image etc.

## References:

Bishop, C. M. (1995). Training with Noise is Equivalent to Tikhonov Regularization. In *Neural Computation* (Vol. 7, Issue 1). http://research.microsoft.com/~cmbishop

Dalal, N., & Triggs, B. (n.d.). Histograms of Oriented Gradients for Human Detection. *2005 IEEE Computer Society Conference on Computer Vision and Pattern Recognition (CVPR'05)*. https://doi.org/10.1109/cvpr.2005.177

Han, S., Liu, X., Mao, H., Pu, J., Pedram, A., Horowitz, M. A., & Dally, W. J. (2016). EIE: efficient inference engine on compressed deep neural network. *ACM SIGARCH Computer Architecture News*, *44*(3), 243–254.

Huang, J.-T., Li, J., Yu, D., Deng, L., & Gong, Y. (2013). Cross-language knowledge transfer using multilingual deep neural network with shared hidden layers. *2013 IEEE International Conference on Acoustics, Speech and Signal Processing*, 7304–7308.

Jiang, Y., Zur, R. M., Pesce, L. L., & Drukker, K. (2009). A study of the effect of noise injection on the training of artificial neural networks. *2009 International Joint Conference on Neural Networks*, 1428–1432.

Maaten, L., Chen, M., Tyree, S., & Weinberger, K. (2013). Learning with marginalized corrupted features. *International Conference on Machine Learning*, 410–418.

Matsuoka, K. (1992). *Noise Injection into Back-Propagation* (Vol. 22, Issue 3).

Murray, A. F., & Edwards, P. J. (1994). Enhanced MLP performance and fault tolerance resulting from synaptic weight noise during training. *IEEE Transactions on Neural Networks*, *5*(5), 792–802.

Reed, R., Marks, R. J., & Oh, S. (1995). Similarities of error regularization, sigmoid gain scaling, target smoothing, and training with jitter. *IEEE Transactions on Neural Networks*, *6*(3), 529–538.

Simard, P. Y., Steinkraus, D., Platt, J. C., & others. (2003). Best practices for convolutional neural networks applied to visual document analysis. *Icdar*, *3*(2003).

Thung, K. H., & Wee, C. Y. (2018). A brief review on multi-task learning. *Multimedia Tools and Applications*, *77*(22), 29705–29725. https://doi.org/10.1007/s11042-018-6463-x

Wang, C., & Principe, J. C. (1999). Training neural networks with additive noise in the desired signal. *IEEE Transactions on Neural Networks*, *10*(6), 1511–1517.

Zhang, J., Ma, C., Liu, J., & Shi, G. (2020). Penetrating the influence of regularizations on neural network based on information bottleneck theory. *Neurocomputing*. https://doi.org/10.1016/j.neucom.2020.02.009

Zhang, Y., & Yang, Q. (2017). A survey on multi-task learning. *ArXiv Preprint ArXiv:1707.08114*.

Zhang, Y., & Yeung, D.-Y. (2014). A regularization approach to learning task relationships in multitask learning. *ACM Transactions on Knowledge Discovery from Data (TKDD)*, *8*(3), 1–31.

Fukushima, K., Miyake, S., & Ito, T. (1983). Neocognitron: A neural network model for a mechanism of visual pattern recognition. *IEEE Transactions on Systems, Man, and Cybernetics*, (5), 826–834.

He, K., Zhang, X., Ren, S., & Sun, J. (2016). Deep residual learning for image recognition. *Proceedings of the IEEE Conference on Computer Vision and Pattern Recognition*, 770–778.

Huang, G., Sun, Y., Liu, Z., Sedra, D., & Weinberger, K. Q. (2016). Deep networks with stochastic depth. *European Conference on Computer Vision*, 646–661.

Hubel, D. H., & Wiesel, T. N. (1968). Receptive fields and functional architecture of monkey striate cortex. *The Journal of Physiology*, *195*(1), 215–243.

Ioffe, S., & Szegedy, C. (2015). *Batch Normalization: Accelerating Deep Network Training by Reducing Internal Covariate Shift*.

LeCun, Y, Boser, B., Denker, J. S., Henderson, D., Howard, R. E., Hubbard, W., & Jackel, L. D. (1998). Backpropagation Applied to Handwritten Zip Code Recognition. *Neural Computation*, *1*(4), 541–551. https://doi.org/10.1162/neco.1989.1.4.541

LeCun, Yann, Boser, B. E., Denker, J. S., Henderson, D., Howard, R. E., Hubbard, W. E., & Jackel, L. D. (1990). Handwritten digit recognition with a back-propagation network. *Advances in Neural Information Processing Systems*, 396–404.

Niu, X.-X., & Suen, C. Y. (2012). A novel hybrid CNN--SVM classifier for recognizing handwritten digits. *Pattern Recognition*, *45*(4), 1318–1325.

Russakovsky, O., Deng, J., Su, H., Krause, J., Satheesh, S., Ma, S., … others. (2015). Imagenet large scale visual recognition challenge. *International Journal of Computer Vision*, *115*(3), 211–252.

Santurkar, S., Tsipras, D., Ilyas, A., & Madry, A. (2018). How does batch normalization help optimization? *Advances in Neural Information Processing Systems*, 2483–2493.

Simonyan, K., & Zisserman, A. (2014). *Very Deep Convolutional Networks for Large-Scale Image Recognition*.

Srivastava, N., Hinton, G., Krizhevsky, A., Sutskever, I., & Salakhutdinov, R. (2014). Dropout: a simple way to prevent neural networks from overfitting. *The Journal of Machine Learning Research*, *15*(1), 1929–1958.

Szegedy, C., Ioffe, S., Vanhoucke, V., & Alemi, A. A. (2017). Inception-v4, inception-resnet and the impact of residual connections on learning. *Thirty-First AAAI Conference on Artificial Intelligence*.

Szegedy, C., Liu, W., Jia, Y., Sermanet, P., Reed, S., Anguelov, D., … Rabinovich, A. (2015). Going deeper with convolutions. *2015 IEEE Conference on Computer Vision and Pattern Recognition (CVPR)*. https://doi.org/10.1109/cvpr.2015.7298594

Zeiler, M. D., Fergus, R., Szegedy, C., Liu, W., Jia, Y., Sermanet, P., … Zisserman, A. (2014). Going deeper with convolutions. *Lecture Notes in Computer Science*, 818–833. https://doi.org/10.1109/cvpr.2015.7298594

Zhang, W., Itoh, K., Tanida, J., & Ichioka, Y. (1990). Parallel distributed processing model with local space-invariant interconnections and its optical architecture. *Applied Optics*, *29*(32), 4790–4797.